\documentclass[letterpaper, 10 pt, conference]{ieeeconf}  % Comment this line out if you need a4paper

\IEEEoverridecommandlockouts                              % This command is only needed if 
\usepackage{cite}
\usepackage{amsmath,amssymb,amsfonts}
\usepackage{algorithmic}
\usepackage{graphicx}
\usepackage{textcomp}
\usepackage{xcolor}
\usepackage{academicons}
\usepackage{siunitx}
\usepackage{booktabs}
\usepackage{multirow}
\usepackage{svg}
\usepackage{fancyhdr}

\usepackage{amsmath}
\newbox{\myorcidaffilbox}
\sbox{\myorcidaffilbox}{\large\includegraphics[height=1.25ex]{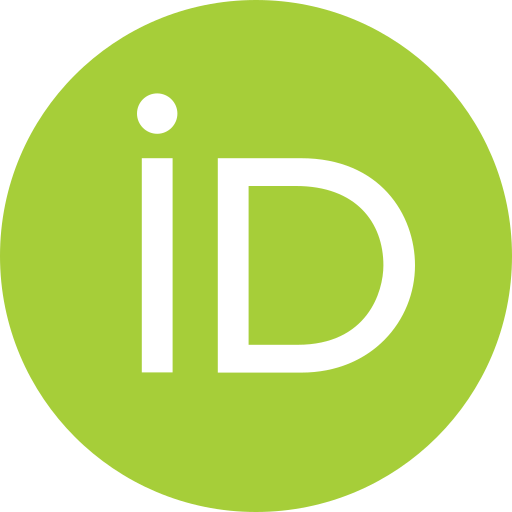}}
\newcommand{\orcidaffil}[1]{%
  \href{https://orcid.org/#1}{\usebox{\myorcidaffilbox}}}

\usepackage{hyperref}
\def\BibTeX{{\rm B\kern-.05em{\sc i\kern-.025em b}\kern-.08em
    T\kern-.1667em\lower.7ex\hbox{E}\kern-.125emX}}

\title{\LARGE \bf Design of a Biomimetic Tactile Sensor \par for Material Classification\\
}

\author{
    Kevin Dai$^{1}$\textsuperscript{\orcidaffil{0000-0002-5895-0450}},
    Xinyu Wang$^{2}$\textsuperscript{\orcidaffil{0000-0003-1058-222X}},
    Allison M. Rojas$^{1}$\textsuperscript{\orcidaffil{0000-0002-5063-0866}},
    Evan Harber$^{2}$\textsuperscript{\orcidaffil{0000-0001-9757-9749}},
    Yu Tian$^{2}$\textsuperscript{\orcidaffil{0000-0001-5919-2519}},
    Nicholas Paiva$^{2}$\textsuperscript{\orcidaffil{0000-0002-5505-8328}},
    \\
    Joseph Gnehm$^{2}$\textsuperscript{\orcidaffil{0000-0003-4212-8304}},
    Evan Schindewolf$^{2}$\textsuperscript{\orcidaffil{0000-0002-8692-8963}},
    Howie Choset$^{2}$$^{*}$,
    Victoria A. Webster-Wood$^{1}$$^{+}$\textsuperscript{\orcidaffil{0000-0001-6638-2687}}, and
    Lu Li$^{2}$$^{**}$\textsuperscript{\orcidaffil{0000-0002-3346-283X}}

\thanks{This work was supported by the CMU Manufacturing Futures Initiative (MFI) and by the National Science Foundation (NSF) Research Fellowship Program under Grant No. DGE1745016. Any opinions, findings, and conclusions or recommendations expressed in this material are those of the authors and do not necessarily reflect the views of the National Science Foundation.}%

\thanks{$^{1}$\textit{Department of Mechanical Engineering, Carnegie Mellon University, 5000 Forbes Ave, Pittsburgh, PA 15213, United States}
}

\thanks{$^{2}$\textit{Robotics Institute, Carnegie Mellon University, 5000 Forbes Ave, Pittsburgh, PA 15213, United States}}

\thanks{$^{*}$\texttt{choset@andrew.cmu.edu}\hfill
$^{+}$\texttt{vwebster@andrew.cmu.edu}}
\thanks{$^{**}$\texttt{lilu12@andrew.cmu.edu}}
}

\begin{document}

\maketitle

\thispagestyle{fancy}
\chead{\fontsize{8pt}{11pt}\selectfont Accepted in 2022 IEEE International Conference on Robotics and Automation (ICRA)}
\cfoot{\fontsize{8pt}{11pt}\selectfont \copyright 2021 IEEE. Personal use of this material is permitted. Permission from IEEE must be obtained for all other uses, in any current or future media, including reprinting/republishing this material for advertising or promotional purposes, creating new collective works, for resale or redistribution to servers or lists, or reuse of any copyrighted component of this work in other works}

\begin{abstract}
    Tactile sensing typically involves active exploration of unknown surfaces and objects, making it especially effective at processing the characteristics of materials and textures. A key property extracted by human tactile perception is surface roughness, which relies on measuring vibratory signals using the multi-layered fingertip structure. Existing robotic systems lack tactile sensors that are able to provide high dynamic sensing ranges, perceive material properties, and maintain a low hardware cost. In this work, we introduce the reference design and fabrication procedure of a miniature and low-cost tactile sensor consisting of a biomimetic cutaneous structure, including the artificial fingerprint, dermis, epidermis, and an embedded magnet-sensor structure which serves as a mechanoreceptor for converting mechanical information to digital signals. The presented sensor is capable of detecting high-resolution magnetic field data through the Hall effect and creating high-dimensional time-frequency domain features for material texture classification. Additionally, we investigate the effects of different superficial sensor fingerprint patterns for classifying materials through both simulation and physical experimentation. After extracting time series and frequency domain features, we assess a k-nearest neighbors classifier for distinguishing between different materials. The results from our experiments show that our biomimetic tactile sensors with fingerprint ridges can classify materials with more than 8\% higher accuracy and lower variability than ridge-less sensors. These results, along with the low cost and customizability of our sensor, demonstrate high potential for lowering the barrier to entry for a wide array of robotic applications, including model-less tactile sensing for texture classification, material inspection, and object recognition. 
\end{abstract}

\section{Introduction}
\begin{figure}[t!]
    \centering
    \includegraphics[width=\linewidth]{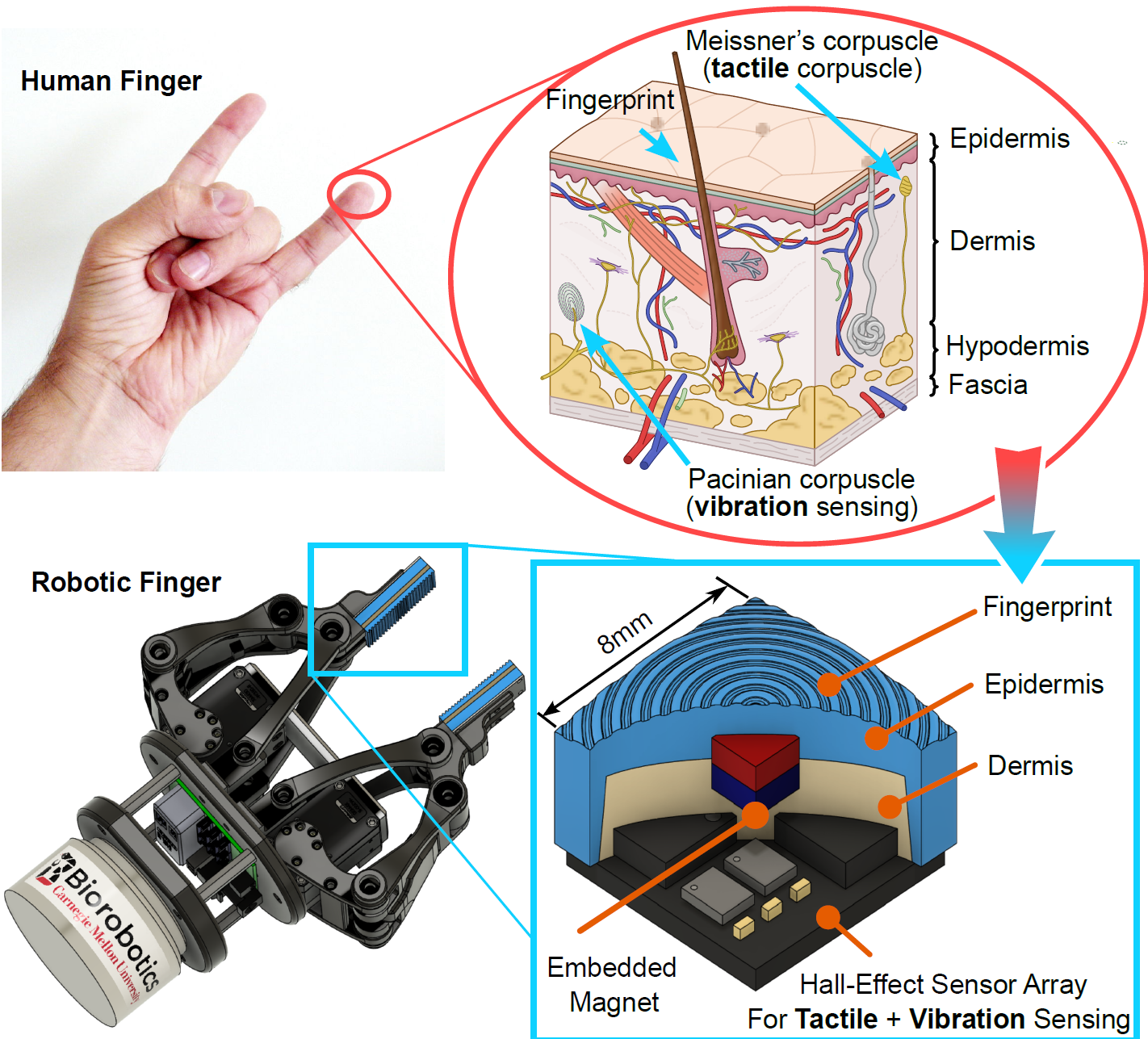}
    \caption{Taking inspiration from human finger tips, we present the design, fabrication, and characterization of a low-cost tactile sensor. Our sensor mimics the layered tissue structure in finger pads, including the dermis, epidermis, and fingerprint. An embedded neodymium magnet and Hall-effect sensor array mimic the function of sensory corpuscles within human fingers. Our sensor's low-cost and customizability make them well-suited for a wide range of tactile sensing applications in robotics. Human hand (Cherus, CC-BY-SA-3.0) and skin structure (Nefronus, CC-BY-SA-4.0) adapted from Wikimedia Commons.}
    \label{sensoroverview}
\end{figure}

Touch, or haptic perception, is one of the five primary senses enabling humans to make observations by interacting with the environment. Tactile sensing provides environmental and material information including surface texture, weight, thermal conductivity, and geometry~\cite{lederman2009haptic}. Human tactile perception is attributed to the geometry and distribution of subcutaneous mechanoreceptors embedded in the epidermis and dermis of skin. One such mechanoreceptor, Pacinian corpuscles (Figure~\ref{sensoroverview}), plays a vital role in fine surface texture perception by transduction of 50-1000 Hz vibrations to neuronal signals~\cite{quintero1984properties, bensmaia2005pacinian, howe1993tactile}.

While humans rely on a complex hierarchical haptic system to understand the physical world, many robots lack this ability, relying instead on cameras and computer-vision approaches. Although vision-based sensors can provide spatial and temporal information about surfaces and objects, low sampling frequencies or visual occlusions during manipulation limit their ability to distinguish surfaces. Furthermore, they are often costly, bulky, and computationally expensive. In scenarios where robots physically manipulate objects, vision-based systems may be insufficient for measuring mechanical properties, whereas touch sensors may lead to an improved performance~\cite{howe1993tactile}. For example, sensors that provide force feedback enable robots to detect slippage when holding objects or reach higher contact precision, increasing the gripping stability~\cite{maeno2000control,howe1989sensing}. Additionally, robots that directly interact with humans may benefit from tactile information for safety by detecting contact with force and torque sensors, adjusting force output based on sensor feedback~\cite{fritzsche2011tactile}. Thus, a wide array of robotic applications would benefit from the availability of a small-scale, low-cost tactile sensor capable of surface characterization, including manufacturing, quality control, and recycling robots as well as robots in human-robot teams.

To address the need for tactile sensation in robots, previous works in tactile sensors have used a variety of technologies such as resistive~\cite{strainGaugeSensor}, optical~\cite{Trueeb_2020}, and capacitive sensing~\cite{CHOU202131}, resulting in sensors with range of functionality, resolution, and cost. Specifically, visuotactile sensors such as GelSight~\cite{johnson2009retrographic,abad2020visuotactile, yuan2015measurement} can provide high resolution texture information but also require complex computational overhead for processing data. Many existing robotic tactile sensors do not have biomimetic features such as mimicking the material properties of skin or utilizing fingerprint ridges. However, for material classification, fingerprints have been demonstrated to play a significant role in amplifying vibrations for transduction \cite{loeb2009role, shao2010finite}. Among the sensors that have implemented biomimetic features, unimodal sensors such as piezoelectric film-based PVDF sensors \cite{yi2017bioinspired} may be relatively easy to manufacture and process data, but may have insufficient degrees of freedom or dynamic range. Multimodal sensors~\cite{fishel2012bayesian} combine many aspects of sensor technology to enable collection of a range of materials' mechanical and thermal properties, and are available in a biomimetic form factor, but require high cost and complexity of fabrication.

A high resolution and high dynamic range but low-cost tactile sensor would lower the barrier to entry for precise determination of applied forces and torques as well as surface texture classification. Recently, magnet-elastomer-based tactile sensors, including work by members of our group \cite{harber2020tunable}, have demonstrated these capabilities in addition to robustness and repeatability while maintaining a small form factor\cite{wang2016design, harber2020tunable}. These tactile sensors embed a magnet in a compliant elastomer and use Hall effect sensors to detect magnetic field gradients corresponding to magnet movement. Multiple degrees of freedom enable the user to determine contact direction, force, and torque in multiple dimensions.

In this work, we extend the sensing capabilities of magnet-elastomer-based tactile sensors by adding biomimetic features, including fingerprint ridges, and evaluate their effect on material texture classification. We hypothesized that magnet-elastomer-based tactile sensors with artificial fingerprint ridges will be able to classify material texture with higher accuracy than sensors without fingerprint ridges.

By utilizing two layers of elastomer with different moduli, we take inspiration from the epidermis and dermis structure of human skin to represent the Pacinian corpuscle mechanoreceptor. In addition, we demonstrate a simple and scalable manufacturing process for customizing tactile sensors. Through both simulation and physical experimentation, we evaluate the material classification capabilities of the sensor with a high-dimensional clustering method. We believe this work expands current capabilities in robotic sensing and makes tactile sensors more accessible, bringing robots one step closer to human levels of tactile perception. 

\begin{figure}[htbp]
\centerline{\includegraphics[width=\linewidth]{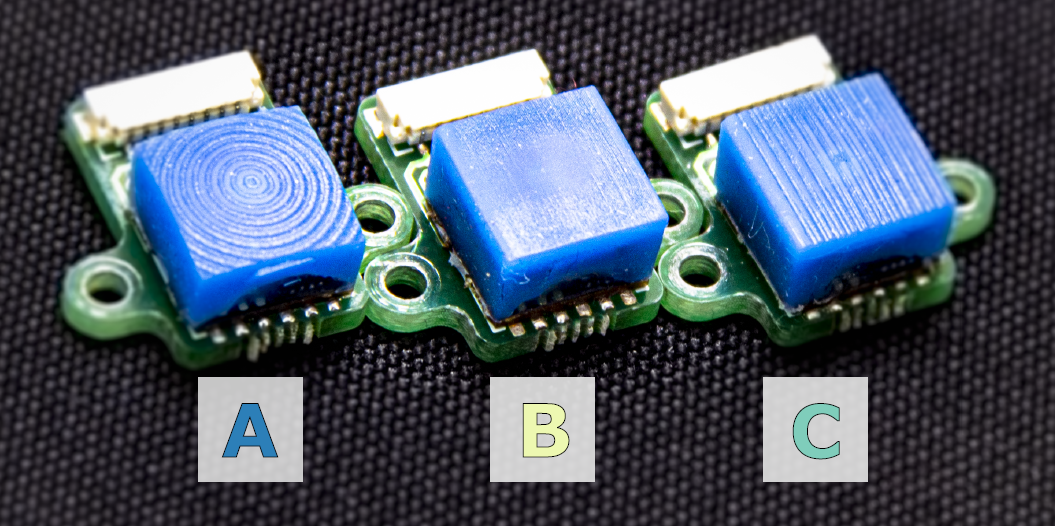}}
\caption{We fabricated three sensor designs to evaluate the hypothesis: \textbf{A} - a spherically curved surface with fingerprint ridges, \textbf{B} - a flat surface without ridges (used as the control group to provide baseline results against designs leveraging artificial fingerprints), and \textbf{C} - a flat surface with fingerprint ridges}
\label{sensordesign}
\end{figure}

\section{Methods}
\subsection{Sensor concept and fabrication}

Our low-cost tactile sensor is designed to use magnetometers (Melexis MLX90393) and a magnet embedded within an elastomer that is attached to the PCB~\cite{harber2020tunable}. When force is applied to the elastomer, the elastomer deforms and moves the magnet, thereby changing the magnetic field detected by the magnetometer through the Hall effect.

The geometry of the tactile sensor's elastomer layers are designed to mimic the structure of a human finger's epidermis and dermis layers. The epidermis layer is 2~mm thick and the dermis layer is 3~mm thick. A 2~mm N50 cube magnet (SuperMagnetMan C0020) is centered between the layers. For some sensor designs (Figure~\ref{sensordesign}), the external surface of the epidermis layer is curved to represent the curvature of human fingers and/or patterned with ridges to represent human fingerprint ridges with a depth of 80~\SI{}{\micro\metre}, width of 400~\SI{}{\micro\metre}, and wavelength of 600~\SI{}{\micro\metre}~\cite{acree1999there, gnanasivam2019gender,mukaibo2005development}.

\begin{figure}[htbp]
\centerline{\includegraphics[width=\linewidth]{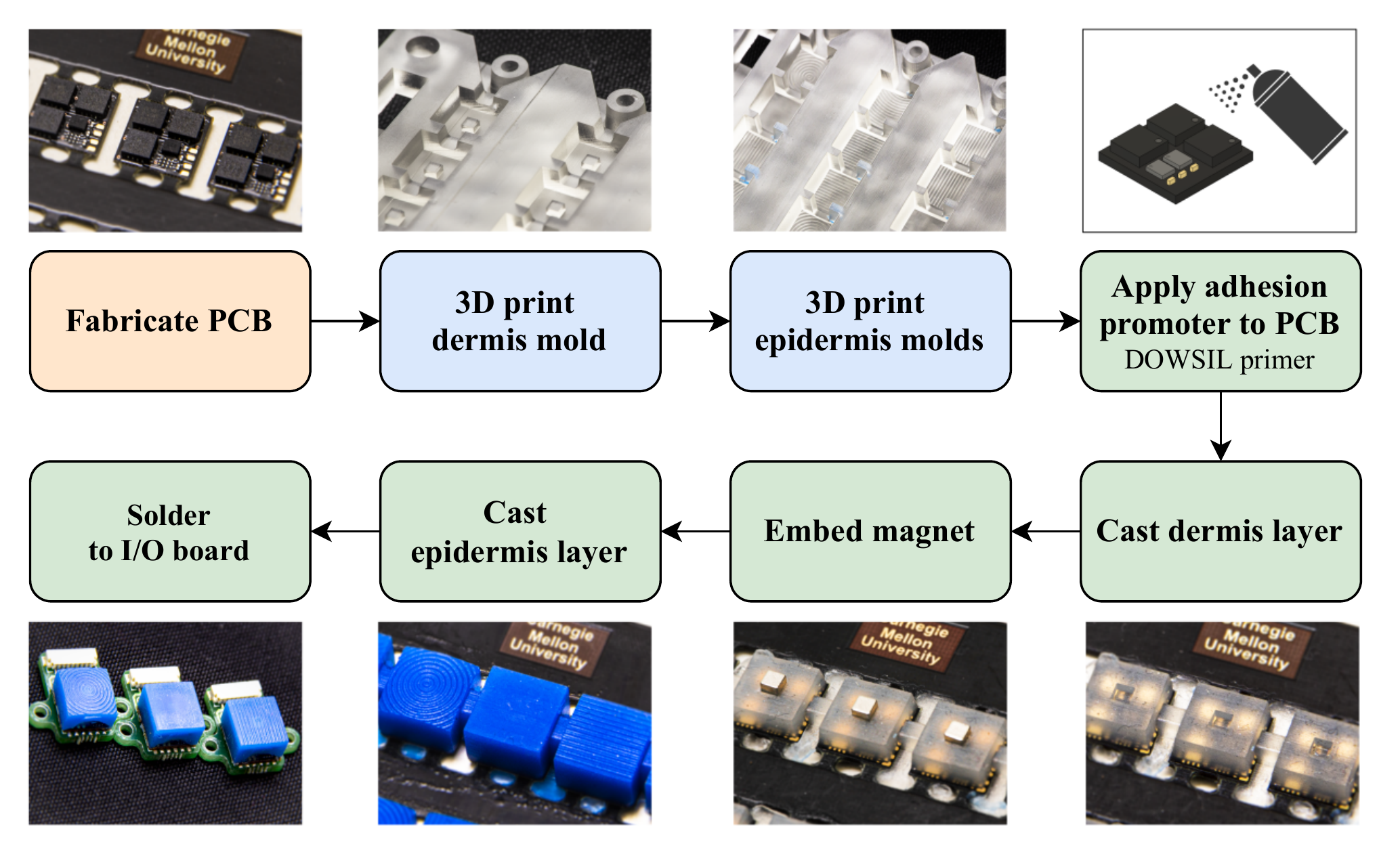}}
\caption{The sensor fabrication process. Sensor PCBs are fabricated as previously described~\cite{harber2020tunable}. A two-stage molding process uses custom 3D printed dermis and epidermis molds. Prior to casting, primer is applied to the PCB, which is then mounted onto the dermis mold for casting the first elastomer layer. After curing, the magnet is embedded in a cavity in the dermis and the second elastomer layer is cast using the epidermis mold. Finally, PCBs are released from the initial board and soldered to the final I/O board.}
\label{sensorfab}
\end{figure}

To fabricate the tactile sensors, we designed a custom mold for injection molding the multiple layers of elastomer. The elastomer was cast onto the PCB to encapsulate the magnetometers and magnet in a two step process for creating the dermis layer and then the epidermis layer. We used stereolithography (Form~2, Clear~v4 resin) to 3D print two molds for the epidermis and dermis layers before rinsing in isopropyl alcohol for 10 minutes and post-curing at 60$^{\circ}$C for 30 minutes. The molds were mounted to the PCB along with a silicone gasket and a rigid acrylic backing plate to prevent elastomer leakage during casting (Figure~\ref{sensorfab}). The epidermis layer (Young's Modulus, $E \approx 145$ psi~\cite{geerligs2011vitro}) were fabricated using Smooth-On Mold Star 30 ($E=96$ psi) and the dermis ($E \approx 20$ psi~\cite{li2012determining, liang2009biomechanical}) were fabricated using Ecoflex 00-10 ($E = 8$ psi). To adhere the silicone rubber to the sensor PCB, we applied DOWSIL 1200 OS primer to the PCB prior to casting. Magnets were encapsulated in the sensor by placing the magnet in a cavity in the dermis casting prior to casting the epidermis layer. 

\subsection{Prediction of sensor performance through simulation}
\begin{figure}[htbp]
\centerline{\includegraphics[width=\linewidth]{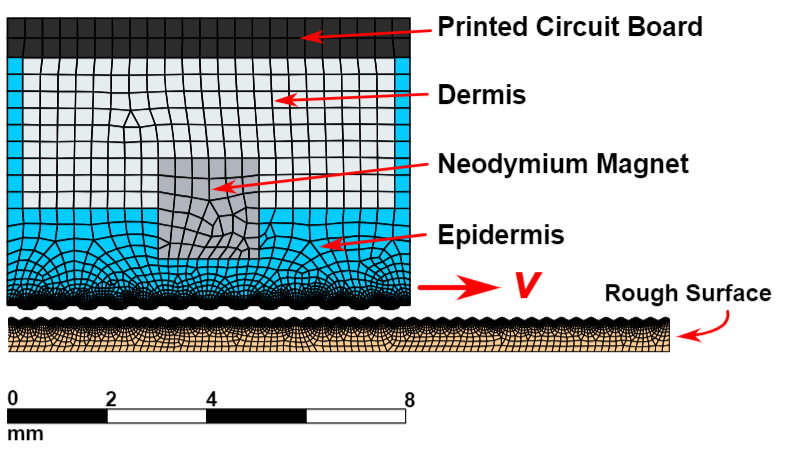}}
\caption{A meshed sensor with fingerprint ridges from the 2D sensor simulation environment in ANSYS. The rough surface is parameterized to allow batch simulation with design points. The neodymium magnet is modeled as a rigid body embedded within the two elastomer layers. Relative movement between the magnet and the PCB is used to calculate triaxial magnetic field values that would be sensed by a magnetometer on the PCB.}
\label{ansysSim}
\end{figure}

To predict the sensor's behavior on a variety of surfaces' roughness and to assess the ability of resulting features to enable surface classification, we created a transient simulation of the tactile sensor. We generated a 2D model of the sensor and a sinusoidal surface using SolidWorks and imported the geometry into ANSYS (Figure~\ref{ansysSim}). Two sensor geometries were represented: (1) a flat sensor without ridges and (2) a flat sensor with parallel fingerprint ridges. We created three batches of configurations by varying contact surface geometry and simulation parameters including sensor velocity, surface amplitude, and surface wavelength (Table~\ref{simulationparam}). The X and Z coordinates of the four corners of the simulated magnet, and outer two corners of the simulated PCB were recorded and exported from ANSYS to CSV files using a custom ANSYS Parametric Design Language (APDL) script.

\begin{table}[htbp]
\caption{Sensor Contact Simulation parameters}
\begin{center}
\begin{tabular}{c c c l}
\toprule
\textbf{Simulation} & \textbf{Parameter}& \textbf{Units} & \textbf{Parameter Values}\\
\midrule
\multirow{3}{5em}{\centering Initial Survey}& Wavelength & mm & 0.06, 0.24, 0.30, 0.60, 5.98\\
&  Amplitude & \SI{}{\micro\metre} & 10, 25, 50, 100\\
& Velocity & mm/s & 25, 50, 100\\
\midrule
\multirow{4}{5em}{\centering Wavelength Sweep} & \multirow{2}{*}{\centering Wavelength} & \multirow{2}{*}{\centering mm} & \multirow{2}{12em}{0.27, 0.33, 0.36, 0.39, 0.42, 0.45, 0.48, 0.51, 0.54, 0.57}\\
&  &  &\\
& Amplitude & \SI{}{\micro\metre} & 10, 25, 50\\
& Velocity & mm/s & 25, 50, 100\\
\midrule
\multirow{3}{5em}{\centering Amplitude Sweep}& Wavelength & mm & 0.24, 0.30, 0.60\\
& Amplitude & \SI{}{\micro\metre} & 15, 20, 30, 35, 40, 45\\
& Velocity & mm/s & 25, 50, 100\\
\bottomrule
\end{tabular}
\label{simulationparam}
\end{center}
\end{table}

\subsection{Experimental data collection for material classification}
To collect experimental sensor data for material classification, we constructed a 2-axis Cartesian robot to repeatably mimic the motion in our design simulations. A 250~mm-long horizontal linear stage driven by a stepper motor formed the X axis. The X axis was mounted to two parallel, 75~mm-long linear stages that were oriented vertically to form the Z axis. The tactile sensor was mounted to the X axis carriage. A TB67S128FTG stepper driver with 128 microstepping was used to drive the Z axis based on step and direction commands. The X axis was driven by a TMC5160 stepper driver with 256 microstepping through SPI using the TMC5160's built-in motion ramp generator. The resolution of Z axis control was 0.4~\SI{}{\micro\metre}, and the resolution of X axis control was 0.2~\SI{}{\micro\metre}. A US Digital EM2 encoder and 2000 LPI encoder strip was used to measure X axis displacements with a resolution of 3.2~\SI{}{\micro\metre}, and an RLS LM13 encoder with 1.0~\SI{}{\micro\metre} resolution was used to measure Z axis displacements. Hall effect sensor data from the tactile sensor and position data from the encoders was sent to a Teensy 4.1 microcontroller, and subsequently recorded over 1000 Hz serial communication to a PC using PuTTY 0.70. Although 3 MLX90393 magnetometer chips were attached to the tactile sensor's PCB, we collected data from a single MLX90393 chip to improve our data collection frequency to 340~Hz. The MLX90393 chips were configured with a conversion ratio of approximately 1.0~\SI{}{\micro\tesla}/LSB.

\begin{figure}[htbp]
\vspace{4pt}
\centerline{\includegraphics[width=\linewidth]{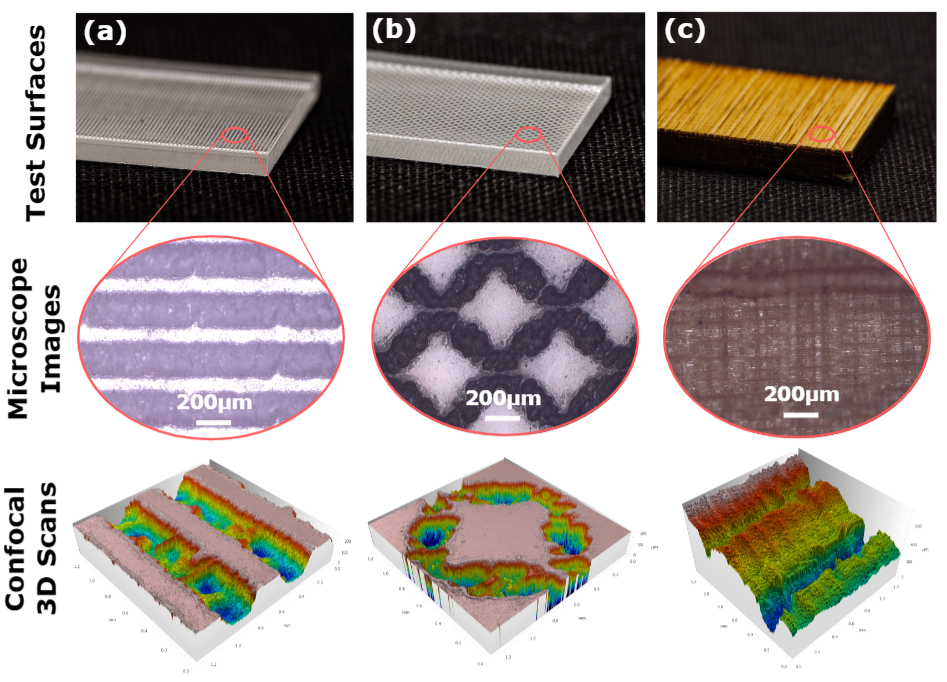}}
\caption{Three examples of materials used in the experimental sensor tests. (\textbf{a}) acrylic etched with lines spaced 0.5 mm apart, (\textbf{b}) acrylic etched with a cross-hatch pattern spaced 0.9 mm apart, (\textbf{c}) wood roughly etched with a stochastically chosen profile. Confocal laser scanned images were used to extract surface roughness parameters from each material.}
\label{material}
\end{figure}

To test our hypothesis that magnet-elastomer-based tactile sensors with ridges will be able to classify material texture with higher accuracy than sensors without fingerprint ridges, each sensor was tested on six material samples of different roughness (Figure~\ref{material}): three acrylic samples, two wood samples, and one aluminum sample. The three acrylic samples and one wood sample were etched with different patterns using an Epilog Helix laser. Each sample measured approximately 190 $\times$ 40 $\times$ 3 mm and was mounted to a rigid acrylic plate on the base of the gantry. We used a Zeiss LSM800 confocal microscope with Zeiss ConfoMap to characterize the surface roughness of each material prior to sensor testing (Table~\ref{materialtable}).

\begin{table}[htbp]
\caption{Material Sample Surface Roughness Characteristics}
\label{materialtable}
\begin{center}
\begin{tabular}{c c c c c} 
\toprule
 \textbf{Material} & \textbf{Description} & \textbf{Ra(\SI{}{\micro\metre})} & \textbf{Rt (\SI{}{\micro\metre})} & \textbf{Rp (\SI{}{\micro\metre})} \\ [0.5ex] 
\midrule
 Aluminum  & Un-etched & 0.795 & 9.37 & 3.68 \\ 
 Wood  & Un-etched & 5.74 & 45.3 & 14.6\\
 Acrylic  & Lines, 1.3\SI{}{\milli\metre} & 7.04 & 110 & 18.7 \\
 Acrylic  & Lines, 0.5\SI{}{\milli\metre} & 10.9 & 71.6 & 31.1 \\
 Acrylic  & Cross-hatched & 15.3 & 264 & 52.6 \\ 
 Wood  & Roughly-etched & 23.8 & 180 & 51.5 \\
 \bottomrule
\end{tabular}
\end{center}
\end{table}

Two sensor designs were selected for experimentation based on our simulation results: (1) flat finger tip without ridges, (2) flat finger tip with parallel ridges. Additionally, we tested a third design inspired by human finger tip geometry that was not captured in the 2D simulation: (3) a spherical finger tip with ridges. During each trial, the sensor was homed to identify the surface contact Z position. Surface contact was determined by calculating the exponential moving average ($\alpha = 0.12$) of the sensor's Z axis magnetic field reading. A contact detection threshold of 10 LSB was set after determining that it corresponded to a contact force of $1.53 \pm 0.41$ g. For each trial the sensor was tested for 3 consecutive repetitions in both positive and negative X-direction translation at 25 mm/s, 50 mm/s, 100 mm/s and 150 mm/s to mimic human exploratory surface scanning speeds~\cite{dahiya2010probing}. Between trials, we changed the material and/or the sensor. We repeated this process over 3 trials, for a total of 3 sensors $\times$ 6 materials $\times$ 4 velocities $\times$ 2 directions $\times$ 3 repetitions $\times$ 3 trials = 1296 data points. Using this data, we determined the sensor's repeatability within each trial as well as across trials after re-positioning the material and sensor relative to each other. 

\subsection{Data processing}
The simulation data was processed to extract sensor data features for use in surface classification based on surface parameters such as wavelength and amplitude. From the simulations' coordinate data recorded in the CSV files, we calculated the displacement and angular rotation of the magnet. We then calculated the 3-axis magnetic field time series data that would be seen by an MLX90393 chip positioned on the PCB using MATLAB~\cite{rojas2021bmes}. The simulated magnetic fields, which had been simulated at a minimum sampling frequency of 5000~Hz, were resampled to a consistent 5000~Hz and then downsampled to 330~Hz for a similar data output frequency as the experimental sensor. Once the time series data of the simulated magnetic field was calculated, simulations and experiments followed the same data processing timeline.

Both experimental data from the gantry and calculated data from the simulations were split based on sensor design before processing both time domain and frequency domain (FFT) data using MATLAB. The experimental time series data was resampled for even spacing in time prior to FFT analysis. A 2 Hz high-pass filter was applied to the data to remove DC effects. From the time series data for each experiment or simulation, we collected statistical features including mean, peak-peak amplitude, standard deviation, skewness, and kurtosis of magnetic field values for each axis. From the FFT, we extracted statistical features including the spectral centroid, standard deviation, skewness, and kurtosis of the power spectrum along both the frequency axis and the power spectral density axis. In addition, we located up to 20 of the largest peaks, if available, with at least 2 prominence in the FFT power spectrum. Then we extracted the statistical features of the 20 peaks along both the frequency axis and the power spectral density axis. Features with the same units were normalized together to a range of~$[0,1]$ using minimum and maximum values.

For evaluating the classification ability of the two simulated sensor designs and three experimentally tested sensor designs, we used the k-nearest neighbors method (k-NN, $k = 5$) with Python. Cross validation was applied with Repeated Stratified K-Fold with 5 folds for a $80\%/20\%$ training/test split. 

\begin{figure*}[htp!]
    \centering
    \includegraphics[width=\linewidth]{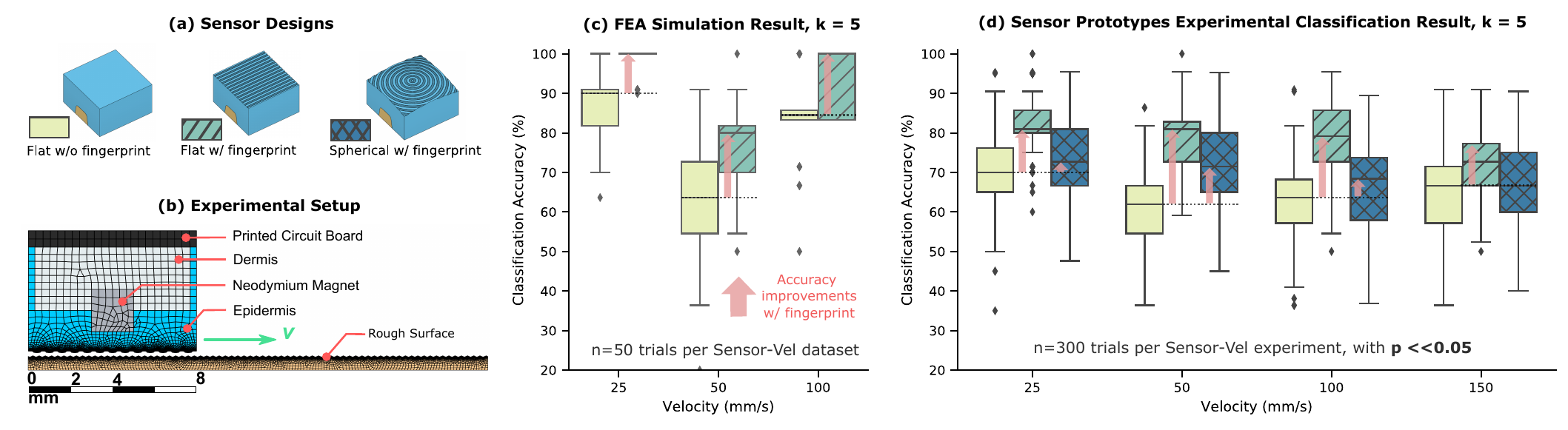}
    \caption{ \textbf{(a)} Three sensor designs used for experimental data collection; the flat surface without ridges acts as the baseline control group to compare against designs that incorporate artificial fingerprints . \textbf{(b)} Illustrated mesh of the ANSYS simulation which mimics the exact physical experimental setup. \textbf{(c)} k-NN ($k = 5$) classification accuracy versus scanning velocity (25 mm/s, 50 mm/s, 100 mm/s) for simulated flat surface sensors with and without ridges.  \textbf{(d)} k-NN ($k = 5$) classification accuracy for physical sensor experiments across four scanning velocities (25mm/s, 50 mm/s, 100 mm/s, 150 mm/s). Red arrows indicate improved classification accuracy with the help of fingerprint ridges.}
    \label{experimentresultsCombined}
\end{figure*}

\subsection{Statistical methods}
We performed a power analysis ($\alpha = 0.05, \beta = 0.80$) on preliminary simulation and experimental results from k-NN ($k=5$) using 5 folds with Repeated Stratified K-Fold and 10 repeats for a total of 50 models. This provided the minimum model count for determining statistical significance between sensors in further analyses. For analyses where the experimental data was split by velocity, we created 300 models for each sensor based on power analysis results. To determine statistical significance, we used ANOVA and Tukey post-hoc tests ($\alpha = 0.05$).

\section{Results}
\subsection{Fingerprint ridges improve surface classification in simulated sensors across velocities tested}

Classification of material surface properties based on simulated sensor data using the k-nearest neighbors method was significantly improved ($p<<0.05$) when sensors included biomimetic fingerprint ridges (Figure \ref{experimentresultsCombined}c). For the three translation velocities simulated, ridged sensors classified surface wavelengths $>8\%$ more accurately than flat, ridge-less sensors. The largest gap in performance was observed at a velocity of 25 mm/s where the ridged sensor resulted in a classification accuracy of $98.1 \pm 3.8\%$, whereas the flat sensor had a classification accuracy of $85.4 \pm 7.9\%$.

\subsection{Fingerprint ridges improve material classification at lower scanning velocities}
To evaluate the effect of scanning velocity on classification accuracy, the experimental results were split by scanning velocity and assessed using the k-nearest neighbors method (Figure~\ref{experimentresultsCombined}d). Classification accuracy was highest for the flat sensor with fingerprint ridges at all velocities ($p<0.05$), which demonstrated the highest accuracy of $82.7 \pm 6.8\%$ at a velocity of 25~mm/s. The addition of curvature to the sensor surface was observed to significantly decrease accuracy across these velocities relative to the flat sensor with ridges. The flat sensor without ridges resulted in the lowest classification accuracies at all velocities, although the observed difference in accuracy between the flat sensor without ridges and the flat sensor with ridges decreased from a gap of $17.8\%$ at a velocity of 50~mm/s to a gap of $7.7\%$ at a velocity of 150~mm/s.

\subsection{Fingerprint ridges improve material classification without prior knowledge of velocity}
\begin{figure}[htbp!]
    \centering
    \includegraphics[width=\linewidth]{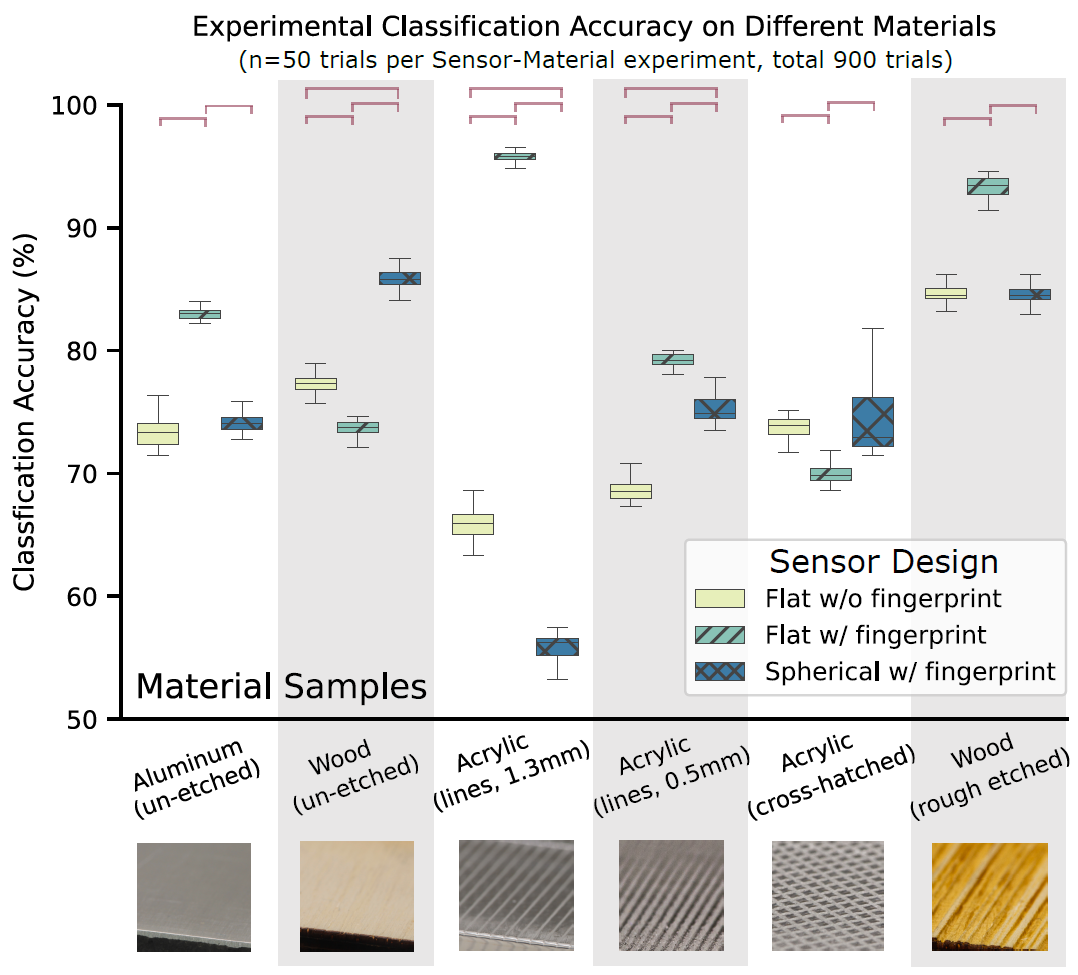}
    \caption{k-NN classification (k = 5) accuracy of each sensor against different material samples (in order of increasing roughness from left to right columns) with outliers removed for better visualization. Significance bars indicate comparisons where $p<0.05$. These experiments are conducted assuming no prior knowledge about scanning velocities, which is considered extremely challenging for material classification tasks using a low-cost tactile sensor without any prior ground truth sensor calibration. }
    \label{material_accuracy}
\end{figure}

The ability of the three sensor designs to accurately classify samples of different materials was assessed using the k-nearest neighbors method without splitting results by velocity (Figure~\ref{material_accuracy}). The classification accuracy for the flat sensor with fingerprint ridges was significantly higher ($p<<0.05$) than the other sensors for the un-etched aluminum, roughly-etched wood, and two acrylic samples with etched lines. The spherical sensor with ridges demonstrated the highest accuracy for the un-etched wood sample. For the cross-hatched acrylic sample, both the spherical sensor with ridges and the flat sensor without ridges performed similarly, with accuracies of $75.1 \pm 4.7\%$ and $ 73.5\pm 3.8\%$, respectively. The largest gap in classification accuracy between sensors was observed for the acrylic sample with etched lines spaced at $1.3\SI{}{\milli\metre}$, with accuracies of $96.1 \pm 1.2\%$ and $ 57.1\pm 6.2\%$ for the flat sensor with ridges and the spherical sensor with ridges, respectively.

\section{Discussion}
Based on our simulations and experimental results, we have demonstrated that our low-cost, magnet-elastomer-based tactile sensors generated sufficient data to allow material classification with high accuracies. Additionally, we have demonstrated that the addition of biomimetic fingerprint ridges to this sensor improved material classification across a wide range of scanning velocities. 

Our 2D simulations of sensor performance provide a design tool for screening the effect of biomimetic features prior to sensor fabrication. Based on these simulation results, we were encouraged to test the addition of fingerprint ridges to our magnet-elastomer-based tactile sensor experimentally, as they indicated improved sensor performance across the scanning velocities tested. This approach could be used in the future to test additional designs and to expand to 3D sensor simulations. Overall, the simulation classification results showed a higher variance than the experimental results. This may be due to the reduced number of samples available for application of the k-nearest neighbors method to the simulated data.  

From our experimental results, both ridged sensors had higher material classification accuracy than the ridge-less sensor at all scanning velocities. The flat sensor with fingerprint ridges maintained the highest classification accuracy, but the gap in accuracy to other sensors decreased at the highest scanning velocity tested. The difference in accuracy between the flat sensor with ridges and the spherical sensor with ridges suggests that the curved geometry may affect the distribution of contact forces acting on the spherical sensor relative to the flat sensor. It is possible that insufficient contact pressure results in few ridges interacting with the material samples and low vibration sensitivity. Further simulation and experimentation is needed to determine the cause for the sensor's performance difference.

Interestingly, the classification accuracy of both types of sensors with ridges appears to gradually decrease with increasing scanning velocity while the classification accuracy of the ridge-less sensor gradually increased with increasing scanning velocity. A non-linear relationship between classification accuracy  and scanning velocity was observed for the flat sensor both in the experiments and in the simulations, with the lowest classification accuracy being observed at 50 mm/s. As the velocity increased above 50 mm/s, the classification accuracy of the flat sensor began to improve. One possibility is that this may be related to the resonant frequency of the sensor. Additional studies are needed to investigate this effect and perform modal analyses of the sensor designs.

One advantage of our tactile sensor is its independence from modeling for accurate classification results. Compared to previous work, our proposed method is not reliant on engineered models and does not require high resolution conversion between sensor readings and physical units. This allows us to avoid extremely time-consuming sensor calibration procedures against expensive ground truth force sensors. To extend the sensor's independence from exterior knowledge, we evaluated the sensor's ability to classify materials without splitting our results data by velocity. The ridged sensors demonstrated the highest classification accuracy for five material samples, suggesting that the tactile sensor could be used in low-cost robotic applications where encoders and other robotic state sensors may not be accessible. Interestingly, we did not detect a noticeable correlation between sensor accuracy and surface roughness using the Ra metric. In the future, studies with alternative roughness metrics and new material samples may explore the sensors' ability to predict continuous surface roughness parameters in addition to discrete classification.

\subsection{Limitations}
Although our 2D simulation provides a solid foundation for sensor design, a few limitations should be noted. First, the simulation time and storage requirement is greatly increased with decreasing time step size. To simulate the faster scanning velocities, the time step must be reduced substantially. On the computer hardware used in this study, this resulted in being unable to simulate the 150 mm/s scanning velocity. In the future, this model should be further improved to allow simulation of a wider range of scanning velocities and include nonlinear material properties. Second, the use of 2D modeling, though it reduces computational time, limits the type of scanning behaviors and surface geometries that can be simulated. A 3D model could be developed in the future to allow exploration of more complex design spaces. However, these limitations did not interfere with testing our overall hypothesis.

\section{Conclusion}
Developing reliable tactile sensors for material classification can be difficult because of limited degrees of freedom and resolution pertaining to the sensor data. The presented sensor design and fabrication procedures provide a feasible solution by utilizing a low-cost magnet-elastomer structure with dual layers, which is modeled on human subcutaneous anatomy. Furthermore, results from both simulation and real-world experiments suggest that fingerprint ridges significantly improve the sensor's ability to classify materials with varying surface properties across a range of velocities. We also verified that sensors with fingerprint ridges are better at classifying materials even without prior information about the scanning velocities. Further work to improve the tactile sensor could utilize additional magnetic field sensory information, create a tactile array, or tune sensor geometry for higher classification accuracy. We believe that this biomimetic tactile sensor has high potential for a wide array of robotic applications, including texture classification, material inspection, object recognition and sorting. 

\newpage
\bibliographystyle{IEEEtran}
\bibliography{ref}

\end{document}